\documentclass[letterpaper, 10 pt, conference]{ieeeconf}  

\usepackage{graphics}
\usepackage[font=small]{caption}
\usepackage{color, soul}
\usepackage{subcaption}
\usepackage{amsfonts}
\usepackage{amsmath}
\usepackage{amssymb}
\usepackage{mathtools}
\usepackage{diagbox}
\usepackage{siunitx}
\usepackage{subcaption}
\usepackage{hyperref}
\usepackage{tikz}
\usepackage[utf8]{inputenc}

\IEEEoverridecommandlockouts                              

\title{\LARGE \bf Robust In-Hand Manipulation with Extrinsic Contacts}

\author{Boyuan Liang$^{1}$, Kei Ota$^{2}$, Masayoshi Tomizuka$^{1}$ and Devesh K. Jha$^{2}$
\thanks{$^{1}$Boyuan Liang and Masayoshi Tomizuka are with the Department of Mechanical Engineering, University of California, Berkeley, CA. {\tt\small \{liangb, tomizuka\}@berkeley.edu}}%
\thanks{$^{2}$Kei Ota and Devesh K. Jha are with Mitsubishi Electric Research Laboratories, Cambridge, MA. {\tt\small \{ota, jha\}@merl.com}}
}%

\begin{document}


\twocolumn[{%
    \renewcommand\twocolumn[1][]{#1}%
    \maketitle
    \begin{center}
        \centering
        \includegraphics[width=0.95\textwidth]{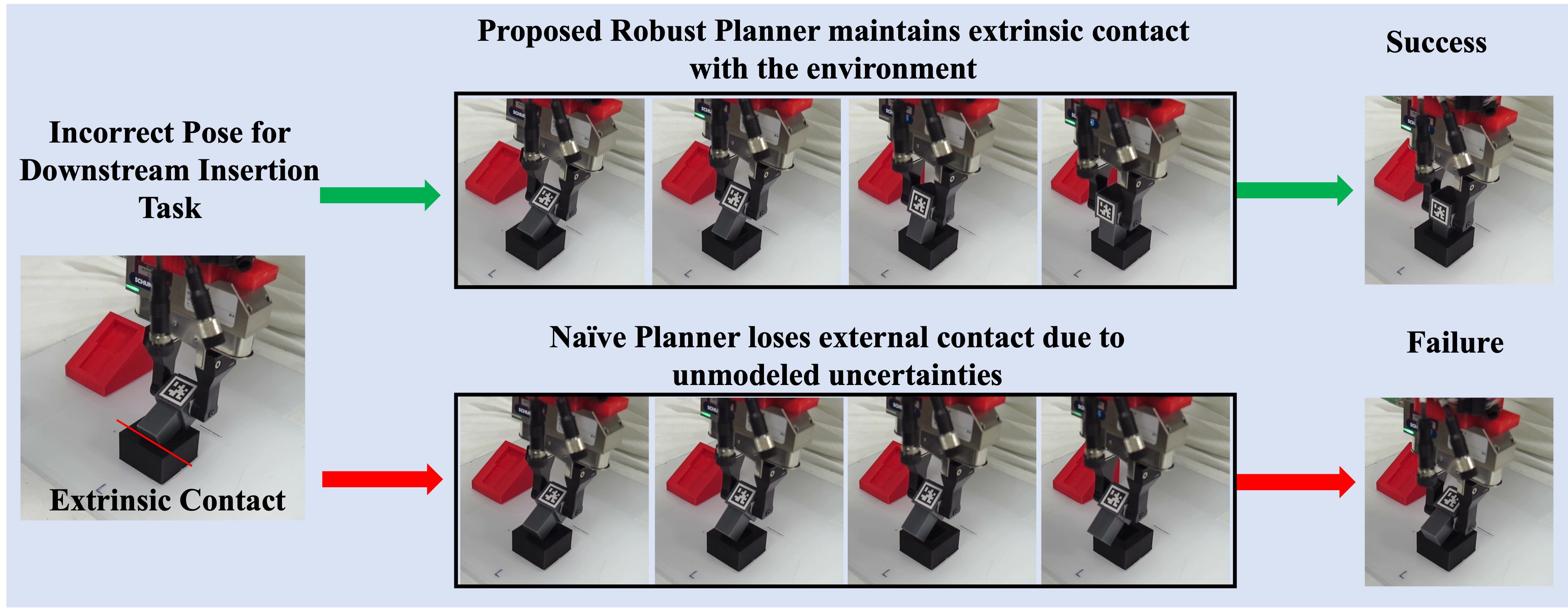} %
    \captionof{figure}{We present in-hand manipulation using an extrinsic line contact (shown by the red line in the initial snapshot). The robot only has a rough estimation of the contact line position. Our proposed robust planning successfully pivots the peg while adjusting its in-hand pose and thus, can successfully perform the desired insertion task. A na\"ive planning approach loses the extrinsic line contact and thus, can not correct the pose of the grasped object. This leads to failure of the plan as seen in the picture.}
    \label{fig:intro-insertion}
    \end{center}%
    }]

\thispagestyle{empty}
\pagestyle{empty}

\footnotetext[1]{Boyuan Liang and Masayoshi Tomizuka are with the Department of Mechanical Engineering, University of California, Berkeley, CA, USA 94720. {\tt\small \{liangb, tomizuka\}@berkeley.edu}}%
\footnotetext[2]{Kei Ota and Devesh K. Jha are with Mitsubishi Electric Research Laboratories (MERL), Cambridge, MA, USA 02139. {\tt\small \{ota, jha\}@merl.com}}


\begin{abstract}
We present in-hand manipulation tasks where a robot moves an object in grasp, maintains its external contact mode with the environment, and adjusts its in-hand pose simultaneously. 
The proposed manipulation task leads to complex contact interactions which can be very susceptible to uncertainties in kinematic and physical parameters.
Therefore, we propose a robust in-hand manipulation method, which consists of two parts. First, an in-gripper mechanics model that computes a na\"ive motion cone assuming all parameters are precise. Then, a robust planning method refines the motion cone to maintain desired contact mode regardless of parametric errors. Real-world experiments were conducted to illustrate the accuracy of the mechanics model and the effectiveness of the robust planning framework in the presence of kinematics parameter errors.

\end{abstract}

\section{Introduction}
\label{sec:intro}

Humans are very skilled at performing dexterous manipulation. We can make very skillful use of various contacts (e.g., with the environment, our own body, etc.) to perform complex manipulation. In a striking contrast, achieving such dexterous behavior for robots remains very challenging. Using environmental contacts efficiently can provide additional dexterity to robots while performing complex manipulation~\cite{mason2018toward}. However, the current generation of robotic systems mostly avoid making contacts with their environment. Contact interactions lead to discontinuous dynamics, and thus, it makes planning, estimation and control of contact-rich tasks involved. Furthermore, generalizable manipulation requires that the planning and control algorithms be robust to uncertainties in kinematic as well as physical parameters of the objects a robot has to manipulate~\cite{cui2021toward}. However, robust planning of manipulation algorithms is generally involved and not well understood~\cite{shirai2022robust, shirai2023covariance}.

In this paper, we study robust in-hand manipulation of objects using extrinsic contacts. Fig. \ref{fig:intro-insertion} shows an example scenario where a robot is holding an object in an incorrect pose for a desired downstream task, and it needs to reorient the object while maintaining the grasp. The proposed in-hand manipulation task presented in the paper can be very susceptible to uncertainties, leading to failures.
Thus, it is desirable that a planning algorithm be robust to various uncertainties like grasp center, extrinsic contact location, etc. We present a method which can incorporate uncertainties in several of the kinematic constraints to generate robust plans for perform in-hand manipulation. This idea is also illustrated in Fig. \ref{fig:intro-insertion}, where a na\"ive plan can easily lose contact with the environment due to uncertainty in the grasp location or the extrinsic contact location.


Maintaining external contact mode is crucial to the success of contact-rich in-hand manipulation tasks (for example see Fig. \ref{fig:intro-insertion}). Several previous works have reported that unexpected contact mode transition can lead to task failure \cite{hou2020manipulation, shirai2022robust}. In this work, a robust in-hand manipulation framework is presented, which fully considers object in-hand slip and parametric uncertainties. We focus on two-finger gripper grasping and scenarios where only translational motion is allowed (see Fig. \ref{fig:intro-insertion}). The proposed robust manipulation framework also finds the gripper motion that minimizes in-hand translation of the object. This is to prevent the robot from losing grasp of the object.


The in-hand manipulation tasks with extrinsic contacts commonly pose the following two major challenges.

\begin{itemize}
    \item \textit{In-hand Motion Modeling}: Given the contact parameters and gripper motion, the system should predict the object motion in both the world and the gripper frame. This is nontrivial since the object is having contact-rich motion in the environment and sliding in the gripper simultaneously. Previous work only solved the situation when the object is static in the environment \cite{chavan2020planar}.
    \item \textit{Robustness Against Parametric Error}: The generated gripper trajectory should prevent unexpected contact mode transition against parametric uncertainties, especially kinematic parameters such as contact position and object dimensions.
\end{itemize}

Consequently, our proposed manipulation framework consists of two parts, respectively corresponding to the two challenges. 1) An \textbf{in-gripper mechanics model} that predicts the object in-hand slip and computes a na\"ive motion cone that maintains desired contact mode assuming environment parameters are precise. 2) A \textbf{robust planning method} that refines the motion cone considering the uncertainty range of each parameter. A trajectory that minimizes in-gripper slip will be generated from the refined motion cone \footnote[1]{A video describing the experiments is available \href{https://youtu.be/YWk4PPY-IE8}{here}.}.

\section{Related Work}
\label{sec:related}

Contact-rich manipulation is the idea to utilize contacts to extend  dexterity of a robotic system\cite{mason2001mechanics}. Several previous work reported its usefulness for non-prehensile manipulation. Without grasping, a single finger can still actuate an object via pivoting \cite{aiyama1993pivoting, aceituno2020global, zhang2023learning, zhang2023simultaneous} and planar pushing \cite{lynch1996stable, hogan2020reactive}. Hou et al \cite{hou2020manipulation, hou2019robust} argued the importance of preventing unexpected contact mode shifts and proposed robust manipulation by maximizing the stable margin of the desired contact mode.

For prehensile manipulation, \cite{dafle2014extrinsic, holladay2015general} showed that even simple grippers can conduct complicated tasks by sequencing motion primitives with external contact. Holladay et al \cite{holladay2015general} modeled gripper contact into a single-point contact and studied pivoting by object gravity and external contact. \cite{chavan2015prehensile, shirai2023tactile} discretized patch gripper contact into multiple point contacts and achieved in-hand manipulation with push primitives. 

Most in-hand manipulation works model gripper-object contact as point contacts. However, for soft or planar fingertips, patched contact is closer to realistic. As a result, some previous work modeled and planned in-hand manipulation using the idea of limit surface (LS) \cite{mason2001mechanics}. It has been reported that ellipsoid LS provides a satisfactory approximation for uniform patched contact \cite{zhou2016convex}. Shi et al \cite{shi2017dynamic} deployed this idea and planned dynamic in-hand manipulation. \cite{chavan2020planar} proposed combining friction cone with ellipsoid LS to generate motion cone for in-hand manipulation.

\section{Mechanics of In-hand Manipulation}
\label{sec:mechanism}
In this section, we present the model for the in-gripper movement of an object in the presence of an external sticking contact. We make the following assumptions for the motion model:


\begin{enumerate}
    \item The object kinematics and physical parameters are known with some uncertainty. 
    \item The parallel-jaw gripper makes a uniform patch contact with the object, so the system can be approximated with a 2D contact model.
    \item The gripper provides enough friction to lift up the object. This assumption is used in subsection \ref{subsec: wrench-motion-set}.
\end{enumerate}

Assume the object has $N$ contact points with the environment, its contact mode is denoted by the following variables in gripper frame of each contact point. a) Contact position $p_i\in\mathbb{R}^2$. b) Contact normal $n_i\in\mathbb{R}^2$. c) Boolean variables $CC_i$, true if the object rotates around the contact point is counter-clockwise. d) Integers $SL_i\in\{-1,0,+1\}$ states if the contact point $i$ is left-sliding, sticking or right sliding.

Let $\mathbf{v}_w$ and $\mathbf{v}_h$ be the object generalized velocity in world frame and gripper frame, respectively. Let $\mathbf{v}_g$ be the gripper motion in world frame. Then, we have $\mathbf{v}_g=\mathbf{v}_w+(-\mathbf{v}_h)$. The mechanics model targets to find a feasible motion cone $\mathbf{\Theta}_g$ of $\mathbf{v}_g$ to maintain the desired contact mode and predict $\mathbf{v}_w$ and $-\mathbf{v}_h$.

Under second-order limit surface (LS) model \cite{shi2017dynamic}, the applicable friction wrench $\mathbf{w}_h=\begin{bmatrix}f_x & f_y & m_z\end{bmatrix}^T$ in gripper frame on the gripper-object patch contact is bounded by an ellipsoid.

\begin{equation}
    \left( \frac{f_x}{\mu_gN_g} \right)^2+\left(\frac{f_y}{\mu_gN_g}\right)^2+\left(\frac{m_z}{\kappa\mu_gN_g}\right)^2\leq 1
\label{eq:model-LS}
\end{equation}

$\mu_g$ and $N_g$ are the coefficient of friction between the object and gripper fingers and the grasping force. $\kappa$ is an integration constant. For uniformly distributed circular patch contact, $\kappa\approx 0.6 r$, where $r$ is the radius of the contact patch \cite{chavan2020planar, shi2017dynamic}. In-hand slip only happens when equality holds in (\ref{eq:model-LS}). Since our manipulation tasks requires in-hand slip, we assume equality always holds in (\ref{eq:model-LS}) \cite{chavan2015prehensile}. By the principle of maximal dissipation, the in-hand motion $\mathbf{v}_h=\begin{bmatrix}v_x & v_y & \omega\end{bmatrix}^T$ should be perpendicular to the LS at the wrench $\mathbf{w}_h$ \cite{chavan2020planar}. This leads to the following parameterized representation of in-hand motion. 
\begin{equation}
\begin{split}
    \mathbf{v}_h&=\alpha\mathbf{w}_h\oslash\tau \\
    \tau&=\begin{bmatrix}(\mu_g N_g)^2 & (\mu_g N_g)^2 & (\kappa\mu_g N_g)^2\end{bmatrix}^T
\end{split}
\label{eq:model-load-motion}
\end{equation}

$\oslash$ stands for element-wise division. $\alpha\geq 0$ is a proportional constant. Given the desired contact mode, the possible range of $-\mathbf{v}_h$ and $\mathbf{v}_w$ are respectively defined as Wrench Motion Set (WMS) and Environmental Motion Set (EMS). The following two subsections explain their computational methods.

\subsection{Wrench Motion Set}
\label{subsec: wrench-motion-set}

(\ref{eq:model-load-motion}) provides a means to infer WMS from the set of all possible $\mathbf{w}_h$, which is named as the wrench set (WS). Linear complementarity conditions of the Coulomb friction model restricts all external contact points can in total provide only one or two wrench rays. We mainly discuss the case when there are two wrench rays, and the case with only one wrench ray can be treated as a degenerated special case. Denote the gravitational wrench in the gripper frame as $\mathbf{G}$ and wrench rays provided from external contact points in the gripper frame as $\mathbf{w}_1$ and $\mathbf{w}_2$, WS can be interpreted as an affine linear cone

\vspace{-0.5em}
\begin{equation}
    WS=\{\mathbf{G}+t_1\mathbf{w}_1+t_2\mathbf{w}_2|t_1,t_2\geq 0\}
\label{eq:model-wrench-cone}
\end{equation}

By assumption 3 (the gripper friction is sufficient to lift up the peg), $\mathbf{G}$ must fall in the LS. As such, intersection between LS and WS is a nonempty 3D ellipse arc in the wrench space, which has the following parameterized representation.

\vspace{-0.5em}
\begin{equation}
    \vec{s}(\theta)=\vec{m}+\vec{a}\cos{\theta}+\vec{b}\sin{\theta}, \quad \theta\in [\theta_l,\theta_h]
\label{eq:model-wrench-arc}
\end{equation}

Combining with (\ref{eq:model-load-motion}), WMS can be represented as

\vspace{-1em}
\begin{equation}
\begin{split}
    &WMS=\alpha(\vec{m}+\vec{a}\cos{\theta}+\vec{b}\sin{\theta})\oslash\tau\\
    &\theta\in [\theta_l,\theta_h], \quad \alpha\geq 0
\end{split}
\label{eq:model-wms}
\end{equation}

\begin{figure}[t!]
\centering
\begin{subfigure}[c]{0.2\textwidth}
\centering
    \frame{\includegraphics[width=\textwidth, trim={0cm 2cm 0cm 2.5cm}, clip]{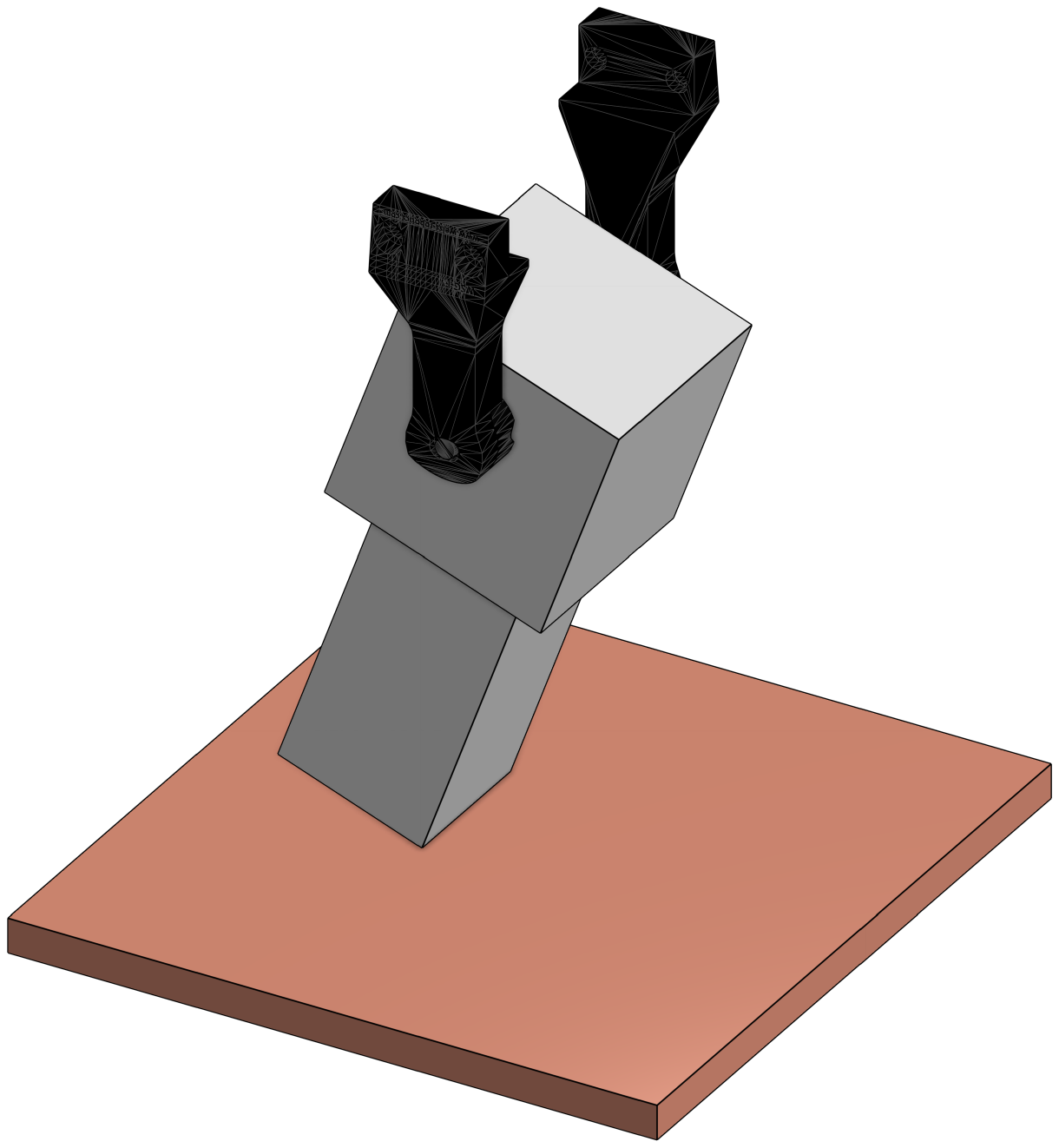}}
\end{subfigure}
\begin{subfigure}[c]{0.07\textwidth}
\centering
    \includegraphics[width=\textwidth]{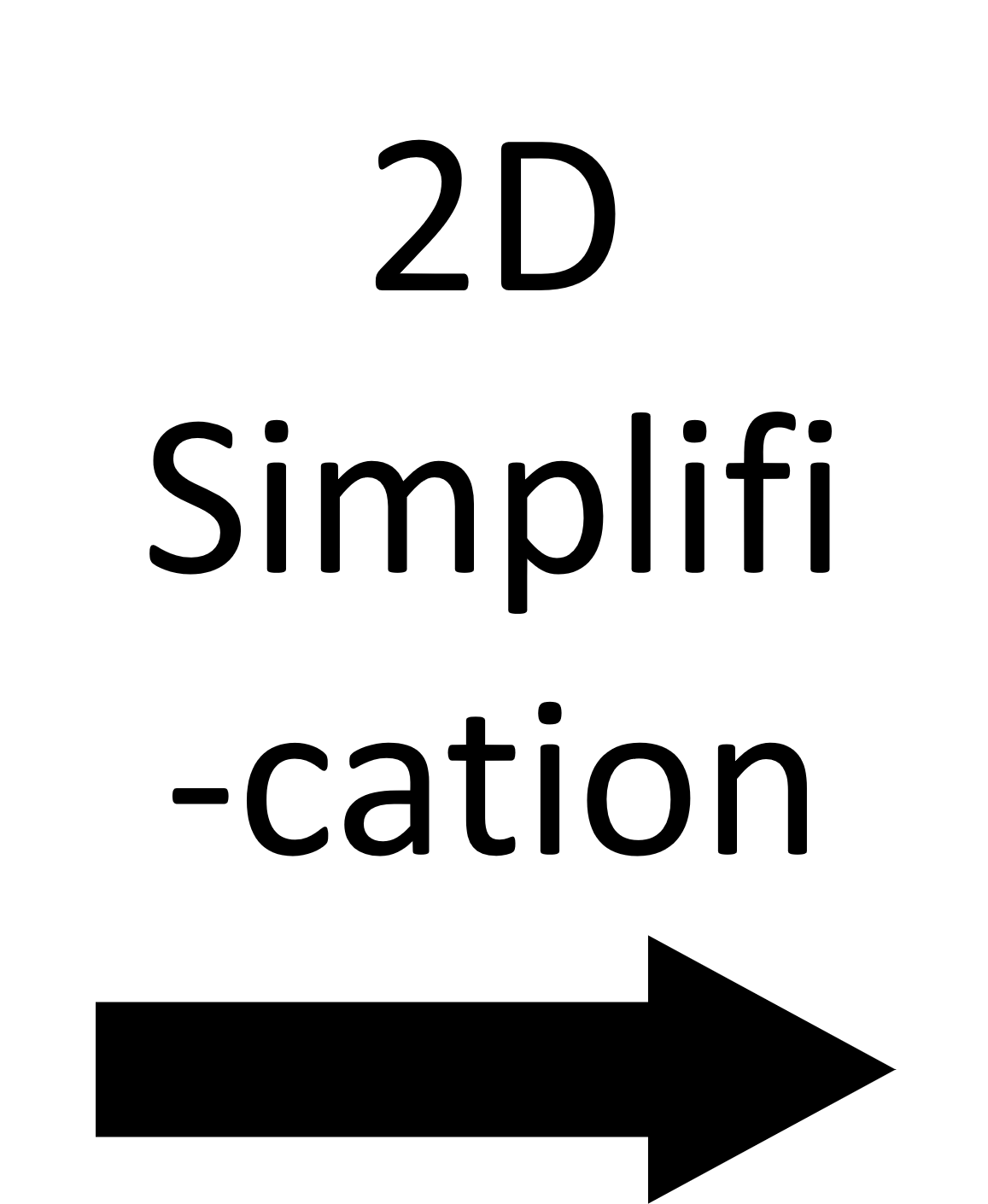}
\end{subfigure}
\begin{subfigure}[c]{0.2\textwidth}
\centering
    \frame{\includegraphics[width=\textwidth, trim={2.5cm 7.5cm 2.5cm 2.5cm}, clip]{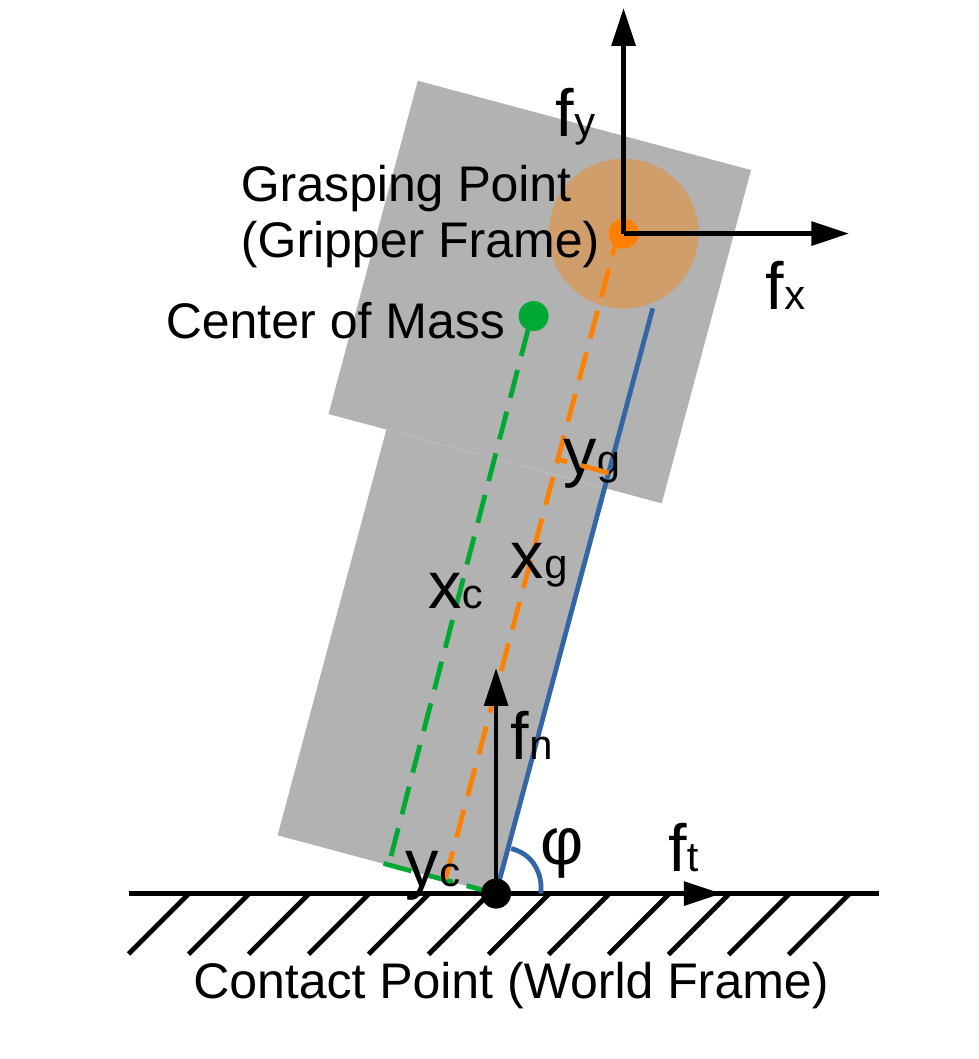}}
\end{subfigure}
\caption{In-hand pivoting with single external line contact. The task is approximated into a 2D contact problem.} 
\label{fig:model_single_pivot_env}
\vspace{-1em}
\end{figure}

\begin{figure}[!t]
\centering
\begin{subfigure}{0.23\textwidth}
    \includegraphics[width=\textwidth, trim=6cm 0.5cm 4cm 4cm, clip=true]{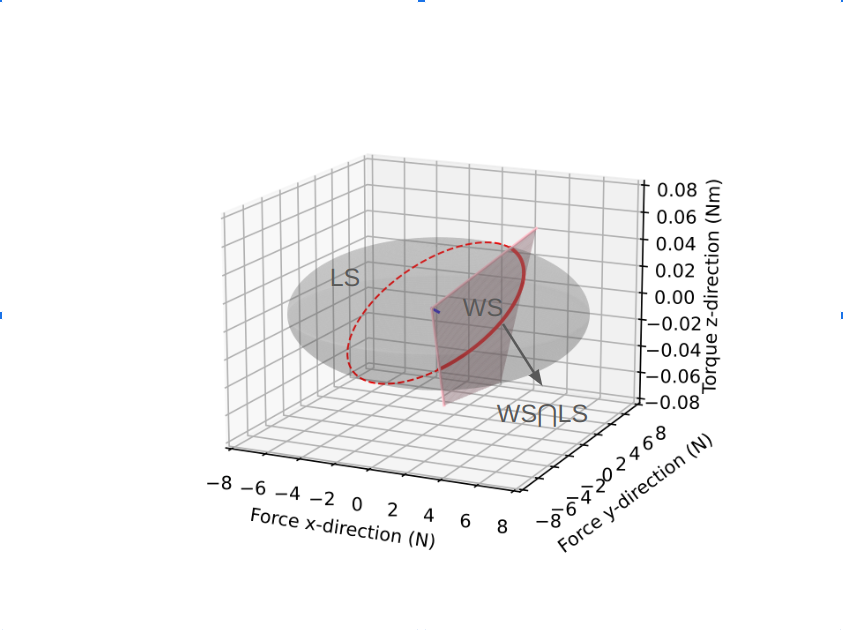}
    \caption{Wrench Space}
    \label{subfig:model-mwcone-normal-ws}
\end{subfigure}
\begin{subfigure}{0.23\textwidth}
    \includegraphics[width=\textwidth, trim=6cm 0.5cm 4cm 4cm, clip=true]{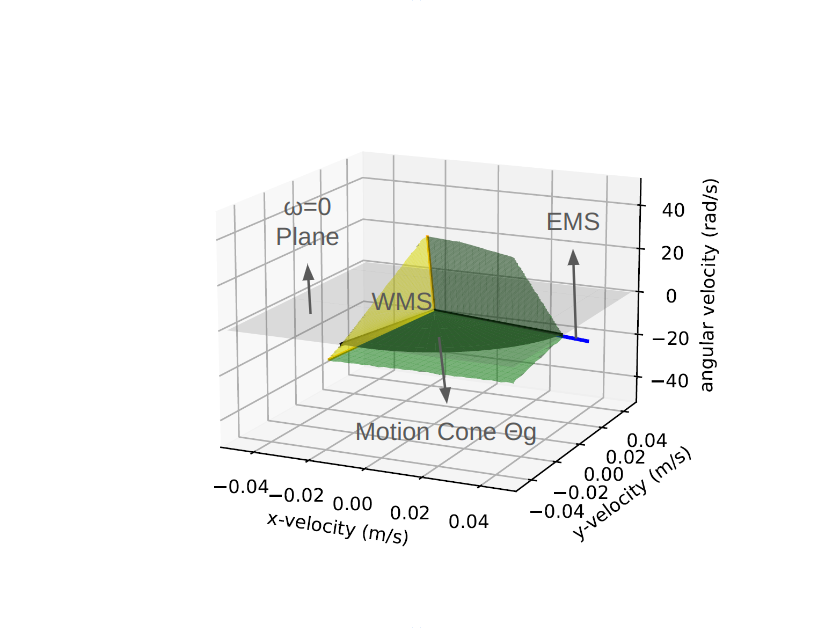}
    \caption{Motion Space}
    \label{subfig:model-mwcone-normal-ms}
\end{subfigure}
\caption{(a) The grey ellipsoid is the LS in (\ref{eq:model-LS}). The two straight pink rays and their spanned wrenches are respectively frictional wrenches $\mathbf{w}_1$, $\mathbf{w}_2$ and $WS$ in (\ref{eq:model-wrench-cone}). The solid red arc is the intersection arc in (\ref{eq:model-wrench-arc}). The short blue line at the origin (so short that is a little hard to see) is the gravitational wrench. (b) WMS and EMS are respectively the yellow surface and the blue ray. The two green plane and the yellow surface enclosed $WMS\bigoplus EMS$, and its intersection with $\omega=0$ plane is $\mathbf{\Theta}_g$, shown by the deep grey wedge.}
\label{fig:model-mwcone-normal}
\vspace{-1em}
\end{figure}

The division in (\ref{eq:model-wms}) is element-wise. We denote the right-hand side of (\ref{eq:model-wms}) by $\alpha\mathbf{v}_{WMS}(\theta)$. Fig. \ref{fig:model-mwcone-normal} is an illustration of WS and WMS taking the case in Fig. \ref{fig:model_single_pivot_env} as example. The robot needs to rotate the peg clockwise around the line contact while maintaining the contact line sticking. System parameters are summarized in the following:

\begin{enumerate}
    \item External coefficient friction $\mu_e=0.25$.
    \item Object mass $m=\SI{0.085}{\kilogram}$. Center of mass location $x_c=\SI{0.041}{\meter}$, $y_c=\SI{0.1}{\meter}$. Directions of $x_c$ and $y_c$ are marked in Fig. \ref{fig:model_single_pivot_env}.
    \item Grasping point location $x_g=\SI{0.01}{\meter}$, $y_g=\SI{0.06}{\meter}$. Directions of $x_g$ and $y_g$ are marked in Fig. \ref{fig:model_single_pivot_env}.
    \item Gripper coefficient of friction $\mu_g=0.4$, patch contact radius $\SI{0.01}{\meter}$, grasping force $N_g=\SI{20}{\newton}$.
    \item Object orientation $\varphi=\ang{70}$.
\end{enumerate}

The intersection between WS and LS are shown in Fig. \ref{subfig:model-mwcone-normal-ws} as the solid red arc. Taking the vertical ray on each point of the solid arc gives WMS, as shown by the transparent yellow surface in Fig. \ref{subfig:model-mwcone-normal-ms}.

\begin{figure}[!t]
\centering
\begin{subfigure}{0.23\textwidth}
    \includegraphics[width=\textwidth, trim=6cm 0.5cm 4cm 4cm, clip=true]{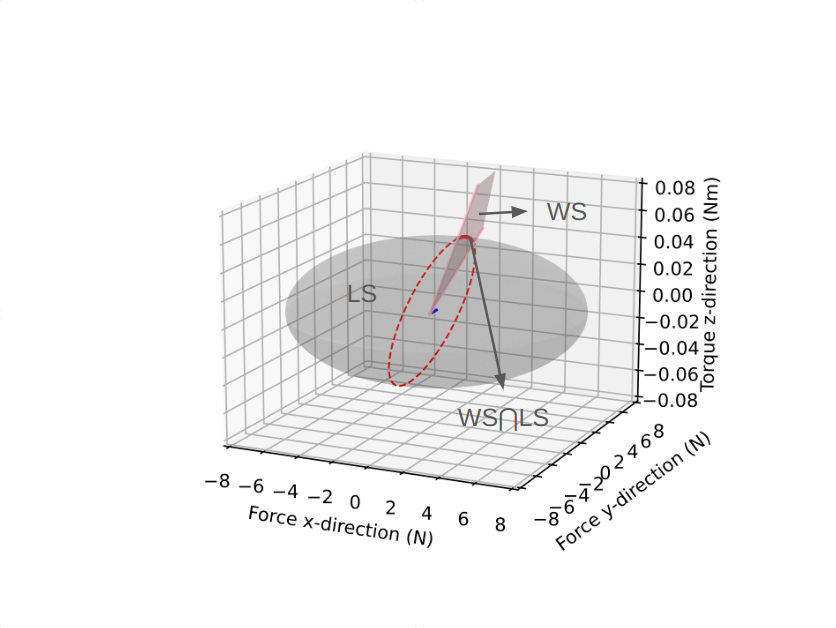}
    \caption{Wrench Space}
\end{subfigure}
\begin{subfigure}{0.23\textwidth}
    \includegraphics[width=\textwidth, trim=6cm 0.5cm 4cm 4cm, clip=true]{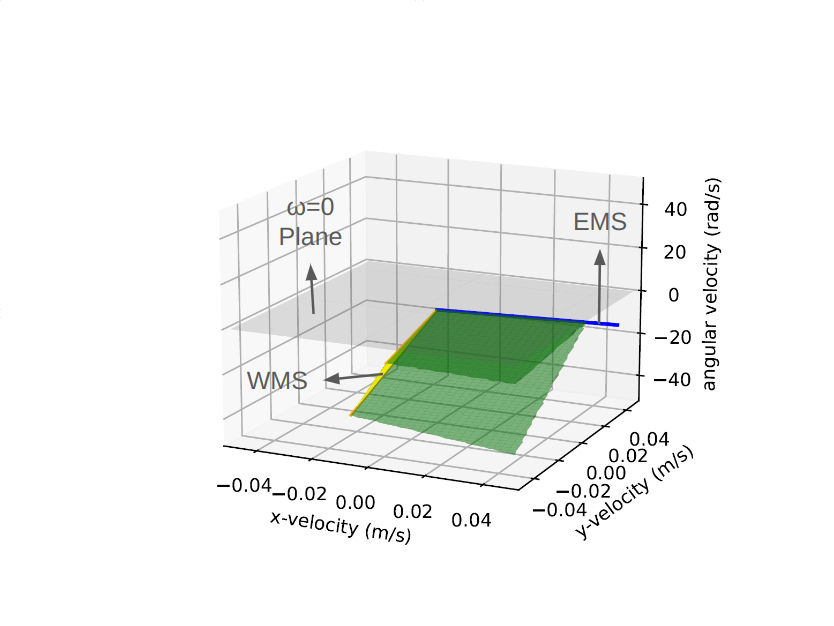}
    \caption{Motion Space}
\end{subfigure}
\caption{Case when $\mathbf{\Theta}_g$ is empty. $WMS\bigoplus EMS$ does not intersect with $\omega=0$ plane, meaning it is infeasible to maintain the desired contact mode.}
\vspace{-1em}
\label{fig:model-mwcone-infeas}
\end{figure}

\begin{figure}[!t]
\centering
\begin{subfigure}{0.23\textwidth}
    \includegraphics[width=\textwidth, trim=6cm 0.5cm 4cm 4cm, clip=true]{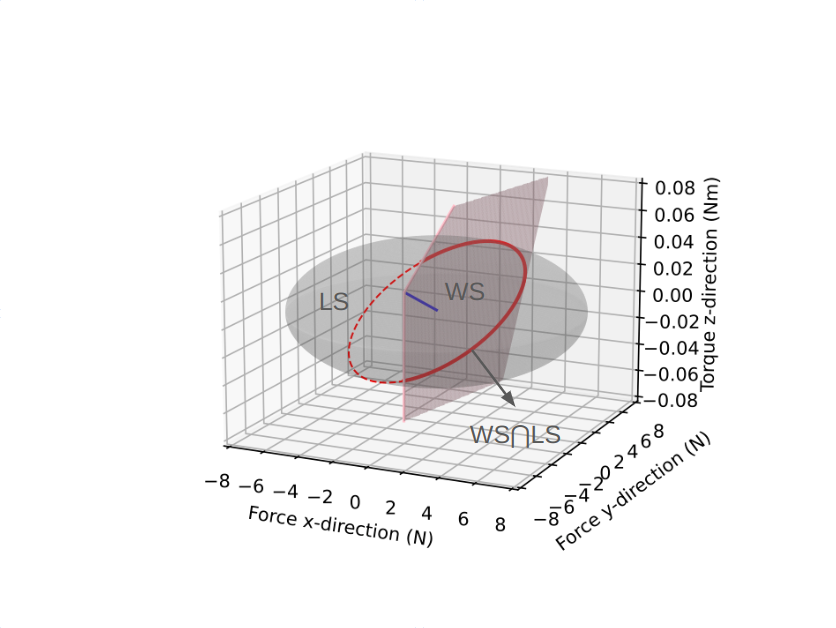}
    \caption{Wrench Space}
\end{subfigure}
\begin{subfigure}{0.23\textwidth}
    \includegraphics[width=\textwidth, trim=4cm 0.5cm 5cm 4cm, clip=true]{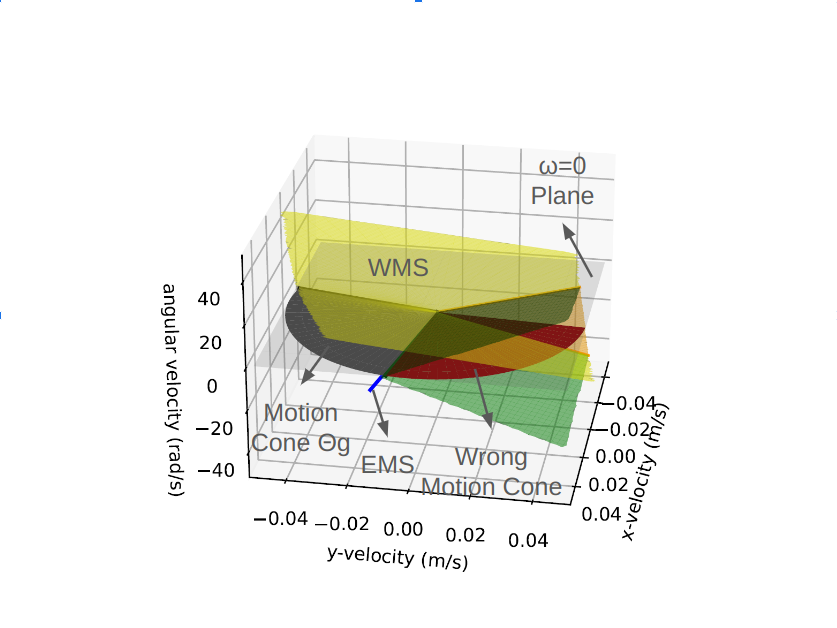}
    \caption{Motion Space}
\end{subfigure}
\caption{Case when $\mathbf{\Theta}_g$ is nonempty and linear cone approximation fails. When (\ref{eq:model-wrench-arc}) is a superior arc, the motion cone generated from linear cone approximation (red wedge) flips to the other side. Most of its elements have positive y-motion, which directly lifts up the object, so the red wedge gives an incorrect motion cone.}
\vspace{-1em}
\label{fig:model-mwcone-fail}
\end{figure}

\begin{figure*}[!t]
\centering
    \includegraphics[width=\textwidth]{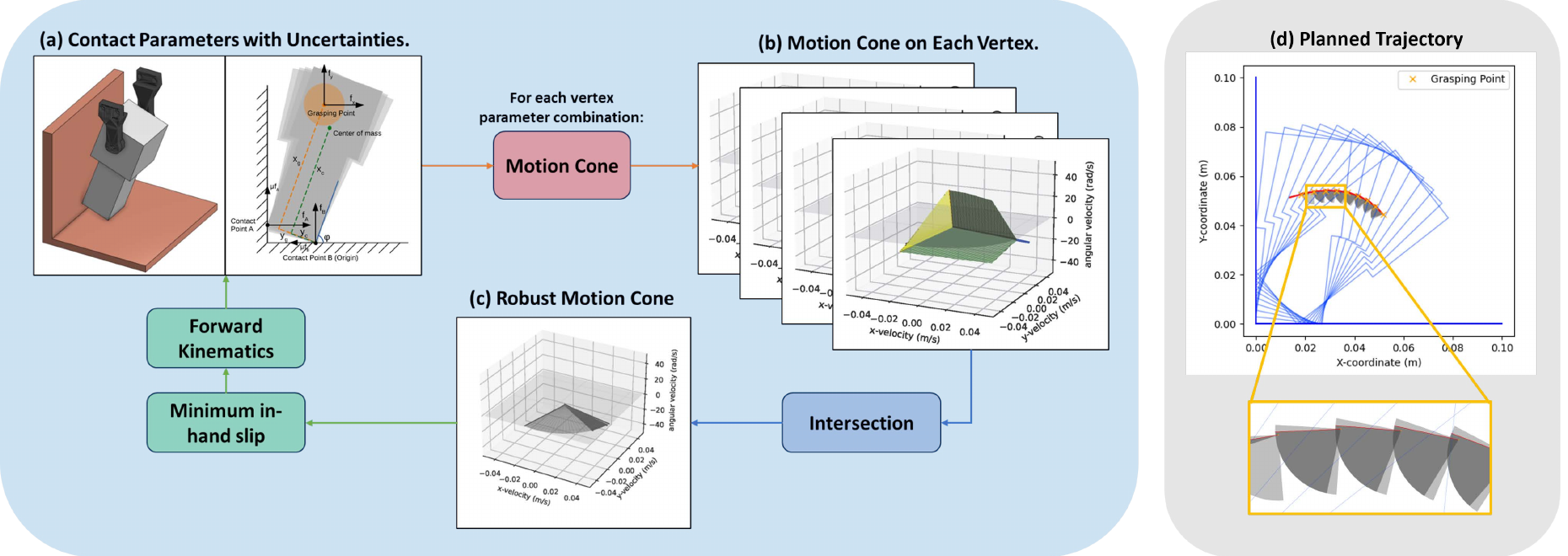}
\caption{Robust Planning Overview. (a) The example scenario where the robot needs to in-gripper pivot the object while maintaining contact with the two external walls. (b) The motion cone generated from each vertex in $\xi\pm\Delta\xi$. (c) The robust motion cone by taking the intersection in (\ref{eq:robust-vertex}), the lighter grey wedge is the na\"ive motion cone assuming parameters are accurate, where the darker grey wedge is the robust motion cone. (d) The planned robust trajectory from $\varphi=\ang{40}$ to $\ang{85}$, the red line is the motion in each step selected from the robust motion cone.} 
\vspace{-1em}
\label{fig:robust-overview}
\end{figure*}

\subsection{Environmental Motion Set}

Computing EMS is straightforward from contact kinematics constraints, which depend on contact mode representation $\{p_i,n_i,CC_i,SL_i\}_{i=1}^N$. For example, the sticking line contact in Fig. \ref{fig:model_single_pivot_env} has contact mode

\[
\begin{split}
    &p_1=\begin{bmatrix}
        x_g\cos{\theta}-y_g\sin{\varphi}\\
        x_g\sin{\theta}+y_g\cos{\varphi}
    \end{bmatrix} \quad
    n_1=\begin{bmatrix}
        0 \\ 1
    \end{bmatrix} \\
    &CC_1=False \quad SL_1=0
\end{split}
\]

It has only one DoF, so the EMS becomes a single ray, which is shown by the blue line in Fig. \ref{subfig:model-mwcone-normal-ms}.

\[
    EMS=\{\beta\begin{bmatrix}
        p_{1,y} & -p_{1,x} & -1
    \end{bmatrix}^T|\beta \geq 0\}
\]

Cases with only one wrench ray has two DoFs, then EMS will become a linear cone spanned by the two motion rays.

\subsection{Motion Cone of Contact Mode}

The feasible motion cone that maintains the desired contact mode is the Minkowski sum $WMS\bigoplus EMS$, which is the cone spanned by the yellow surface and the blue ray in Fig. \ref{subfig:model-mwcone-normal-ms}. Since the gripper is only allowed translational motion, we take its intersection with the $\omega=0$ plane, which forms the motion cone $\mathbf{\Theta}_g$, visualized by the dark grey wedge in Fig. \ref{subfig:model-mwcone-normal-ms}. We quantitatively represent $\mathbf{\Theta}_g$ as an angle range $[\Theta_{g,1},\Theta_{g,2}]$. Given a gripper motion $\mathbf{v}_g$ in the motion cone, it can be uniquely decomposed into a motion in WMS and EMS, which predicts object in-gripper motion.

\begin{equation}
    \mathbf{v}_g=\mathbf{v}_w+(-\mathbf{v}_h), \quad -\mathbf{v}_h\in WMS \quad \mathbf{v}_w\in EMS
\end{equation}

There are also cases when $WMS\bigoplus EMS$ does not intersect with the $\omega=0$ plane. This happens when it is infeasible for the robot to maintain the desired contact mode with only translational motion. Fig. \ref{fig:model-mwcone-infeas} illustrates this case, all system parameters are the same as Fig. \ref{fig:model-mwcone-normal} except
$\varphi=90^\circ$ and $\mu_e=0.1$. Every translational motion will make the contact point slip or the object will remain static due to the small $\mu$, so $\mathbf{\Theta}_g=\emptyset$.

Notice that the WMS, which is the yellow surface in Fig. \ref{subfig:model-mwcone-normal-ms} is not a linear cone. Its curvature becomes more obvious with a larger object mass and environmental coefficient of friction. Previous work approximated WMS as a linear cone spanned by the edges of WMS, which are motion rays taking $\theta=\theta_l$ or $\theta_u$ in  \cite{chavan2020planar}. Linear cone approximation may be valid when (\ref{eq:model-wrench-cone}) is a minor arc like the example in Fig. \ref{fig:model-mwcone-normal}, but Fig. \ref{fig:model-mwcone-fail} illustrates its failure when (\ref{eq:model-wrench-cone}) becomes a superior arc. When the system parameters are the same as Fig. \ref{fig:model-mwcone-normal}, but object mass is increased to $\SI{400}{\gram}$ and $\mu$ is increased to $0.5$, the linearized motion cone flips to the other side. These motions simply lift up the object from the contact plane, which is incorrect. This shows that linear cone approximation is in general not applicable when the peg and gripper are moving together.

\section{Robust Planning Method}
\label{sec:robust}

The motion cone computed in section \ref{sec:mechanism} assumes a precise estimation of each contact parameter. In this section, we refine the motion cone to make it robust against parametric uncertainties. Using $\xi\in\mathbb{R}^n$ to denote the set of parameters whose uncertainties will be considered. For each entry $\xi_i$ of $\xi$, let its uncertainty range be $[\xi_i-\Delta\xi_i,\xi_i+\Delta\xi_i]$, and we use $\Delta\xi$ to denote the collection of $\Delta\xi_i$. Each combination of parameters in the $n$-dimensional cuboid $\Tilde{\xi}\in\xi\pm\Delta\xi$ generates a different motion cone $\mathbf{\Theta}_{g,\Tilde{\xi}}$. The robust motion cone $\bar{\mathbf{\Theta}}_g$ should be the intersection of $\mathbf{\Theta}_{g,\Tilde{\xi}}$ of all possible $\Tilde{\xi}$.

\begin{equation}
    \bar{\Theta}_g=\bigcap_{\Tilde{\xi}\in\xi\pm\Delta\xi}\Theta_{g,\Tilde{\xi}}
\label{eq:robust-strict}
\end{equation}

Practically computing this infinite intersection requires approximation. When $\Delta\xi_i$ is relatively small compared to $\xi_i$, the motion cone $\mathbf{\Theta}_{g,\xi}$ approximately changes monotonically in the uncertainty range with each single parameter $\xi_i$ when other parameters are fixed. Therefore, (\ref{eq:robust-strict}) can be approximated as the intersection of $\mathbf{\Theta}_g$ generated by vertices of the cuboid $\xi\pm\Delta\xi$.

\begin{equation}
[\bar{\Theta}_{g,1},\bar{\Theta}_{g,2}]=:\bar{\mathbf{\Theta}}_g=\bigcap_{\Tilde{\xi}\in\mathcal{V}(\xi\pm\Delta\xi)}\mathbf{\Theta}_{g,\Tilde{\xi}}
\label{eq:robust-vertex}
\end{equation}

With the computed robust motion cone $\bar{\mathbf{\Theta}}_g$, many search-based planning methods such as greedy search and RRT \cite{lavalle2001randomized, wu2020r3t} can be directly deployed. In this work, given fixed angular component in $\mathbf{v}_w$, we focus on finding $\mathbf{v}_g$ at each time step that minimizes the translational part of $-\mathbf{v}_h^{(t)}$. Since non-linear trigonometric constraints can easily trap optimizers in an infeasible local minimum, we apply a sample-based approach. Candidates $\theta$ are uniformly sampled in a prefixed angle range $[\theta_l,\theta_u]$ with common difference $\delta\theta$, and then solve $\alpha\geq 0$, $\beta\geq 0$ and $\Theta\in\bar{\mathbf{\Theta}}_g$ that satisfies the following equality constraints:

\begin{equation}
    \begin{bmatrix}
        \cos{\Theta} & \sin{\Theta} & 0
    \end{bmatrix}^T=\alpha\mathbf{v}_{WMS}(\theta)+\beta\mathbf{v}_w
\end{equation}

The $\theta$ that gives the smallest $\alpha$ will be selected. The solved $\alpha$ and $\beta$ predicts $\mathbf{v}_h$ and $\mathbf{v}_w$, which formulates the forward kinematics.

Fig. \ref{fig:robust-overview} gives an overview of the greedy robust planning method and an example of a planned robust trajectory. There are two external contact points and the robot needs to pivot the peg from $\varphi=\ang{40}$ to $\ang{85}$ while ensuring the two contact points do not separate. System parameters are summarized in the following:

\begin{enumerate}
    \item The horizontal and vertical contact plane has respectively $0.1$ and $0.12$ coefficients of friction.
    \item The grasping point $x_g=\SI{0.055}{\meter}$ and $y_g=\SI{0.0135}{\meter}$, each with a $\pm\SI{3}{\milli\meter}$ uncertainty. Directions of $x_g$ and $y_g$ are marked in Fig. \ref{fig:robust-overview}(a).
    \item The bottom length of the object is $\SI{0.027}{\meter}$, with a $\pm\SI{2}{\milli\meter}$ uncertainty.
    \item Object mass $m=\SI{0.085}{\kilogram}$. Center of mass $x_c=\SI{0.041}{\meter}$, $y_c=\SI{0.1}{\meter}$, each with a $\pm\SI{3}{\milli\meter}$ uncertainty. Directions of $x_c$ and $y_c$ are marked in Fig. \ref{fig:robust-overview}(a).
    \item Gripper coefficient of friction $\mu_g=0.4$, patch contact radius $\SI{0.01}{\meter}$, grasping force $N_g=\SI{20}{\newton}$.
\end{enumerate}

From Fig. \ref{fig:robust-overview}(d) we can see that $\bar{\mathbf{\Theta}}_g$ is a subset of $\mathbf{\Theta}_g$ that further ensures contact mode under parametric uncertainties.

\section{Experiments}
\label{sec:experiments}
In this section, we present results of different experiments performed to verify the effectiveness and correctness of the proposed idea. Two sets of experiments were conducted, each respectively verifying the proposed model in section \ref{sec:mechanism} and showing the effectiveness of the robust planning method in section \ref{sec:robust}. We implemented the proposed framework with a 6 DoF MELFA RV-5AS-D Assista manipulator arm with a Schunk WSG-32 gripper. We put tape on each gripper finger to make the gripper coefficient of friction close to 0.4. An april-tag based vision system \cite{olson2011apriltag} is used to localize and grasp objects during the experiments\footnote[1]{A video describing the experiments is available \href{https://youtu.be/YWk4PPY-IE8}{here}.}.

\subsection{Model Verification}
In the first set of experiments, we verify the correctness of the models for the proposed in-hand manipulation. 
We verify the proposed model's accuracy by checking its indicated feasibility. If the model suggests that a contact mode is infeasible to maintain under the current environmental setting, we try various motions of the gripper to verify the infeasibility.

\begin{figure}[t!]
\centering
\begin{subfigure}{0.22\textwidth}
    \centering
    \includegraphics[width=\textwidth, trim=6cm 0cm 12cm 2cm, clip=true]{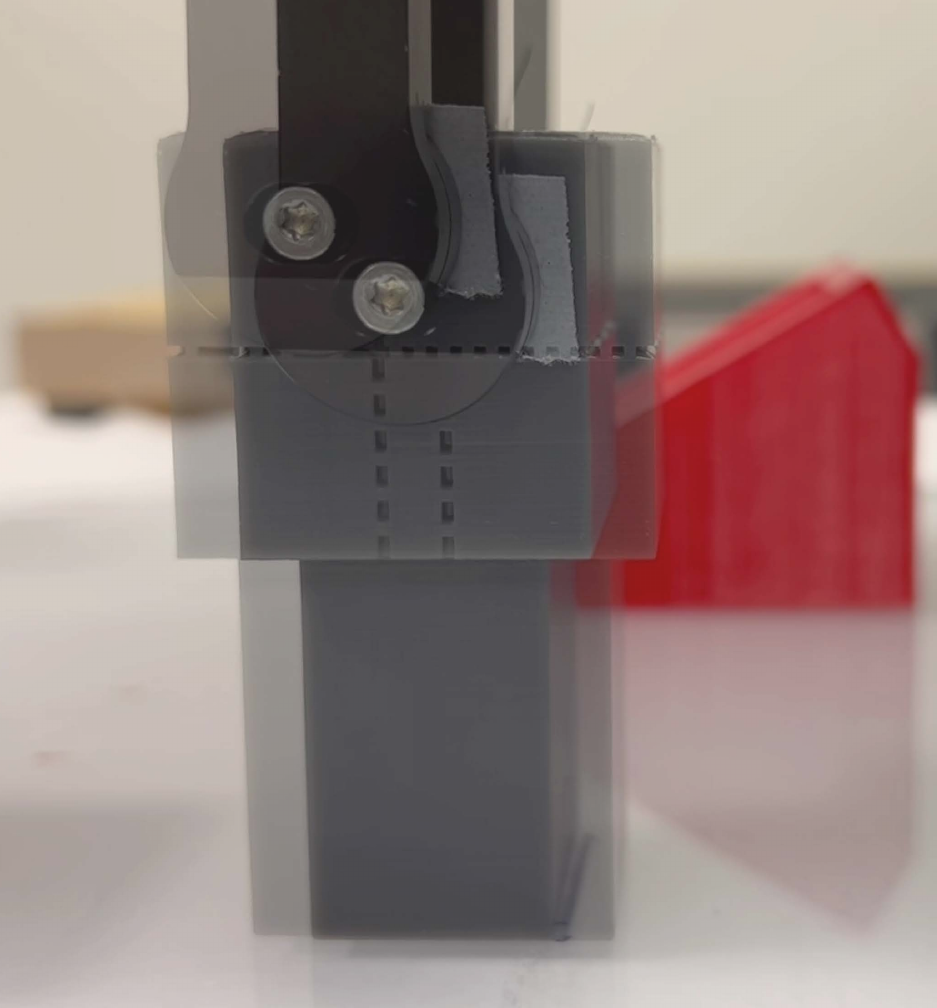}
\end{subfigure}
\begin{subfigure}{0.22\textwidth}
    \centering
    \includegraphics[width=\textwidth, trim=6cm 2cm 12cm 0cm, clip=true]{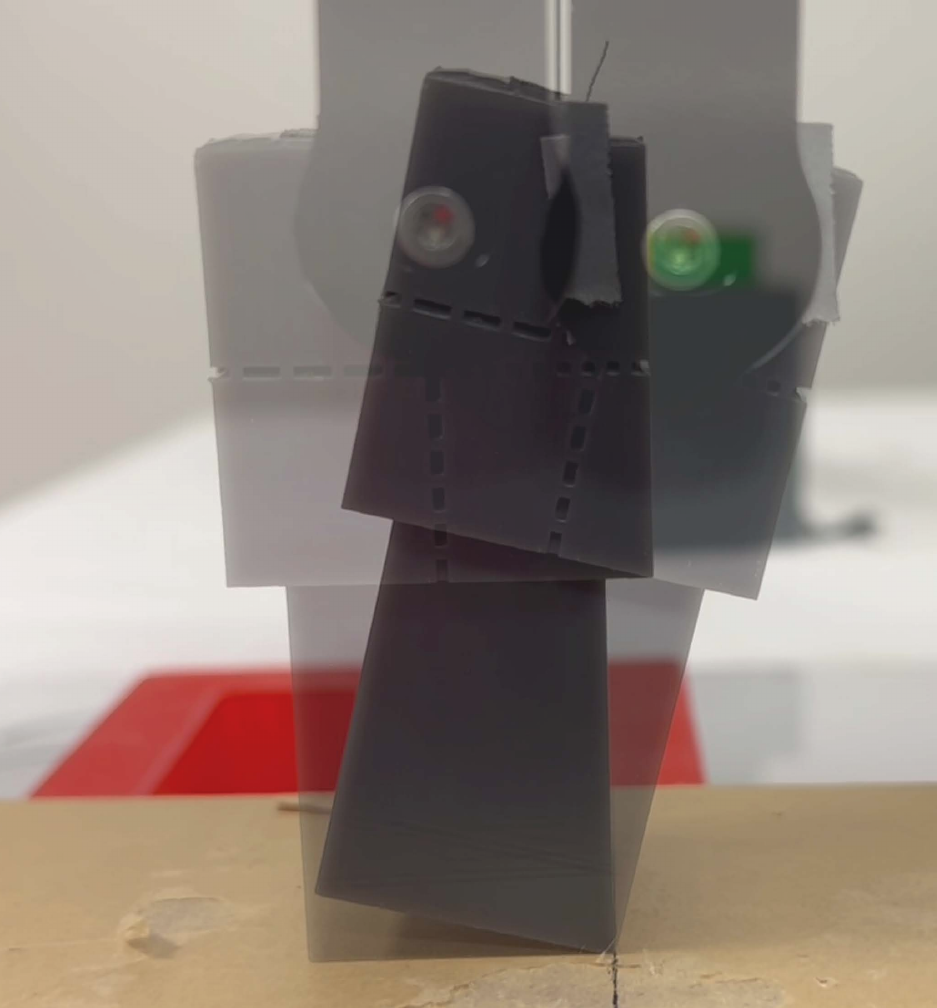}
\end{subfigure}
\caption{Snapshots of single point pivoting experiments. On the glass surface, the contact point keeps sliding, so the robot is unable to pivot it. On the wooden surface, the pivoting succeeded as predicted by the model.}
\vspace{-1em}
\label{fig:exp-verify-a-snapshot}
\end{figure}

The example in Fig. \ref{fig:model_single_pivot_env} is used as the first test case. When the friction coefficient $\mu$ is $0.1$, our proposed model in Section~\ref{sec:mechanism} predicts infeasibility of reorientation while keeping the external contact fixed which is also demonstrated in Fig. \ref{fig:model-mwcone-infeas}. To verify this, we tried to pivot the object on a plexi-glass surface, for which the coefficient of friction is approximately $0.1$. We commanded the gripper to move $1$ cm in ten different directions $\{\cos{\ang{10}i},\sin{\ang{10}i}\}_{i=0}^9$ to see if the peg could be pivoted while maintaining sticking external contact point. The experiment showed that the robot was not able to pivot the object in all ten trial motions. When $\mu=0.25$, a robust trajectory is successfully planned to rotate $\varphi=\ang{90}$ to $\ang{60}$. To verify this, we then switched to a wooden surface where the coefficient of friction is approximately $0.25$. This time, we directly run the planned pivoting trajectory, and the experiment showed that the pivoting was successful. The snapshots of the experiment are presented in Fig. \ref{fig:exp-verify-a-snapshot}.

The second test case is from Fig. \ref{fig:robust-overview}(a). The object's goal pose is $\varphi=\ang{85}$ while ensuring the two contact points do not separate. The proposed model outputs that if the starting pose is $\theta=0$ (or approximately close to $0$), one of the contact points will always separate at the beginning of the planned manipulation. To verify this, we try to pivot the peg with six different gripper motion directions, $\{(\cos{\phi},\sin{\phi})|\phi\in\{\ang{100}, \ang{120}, \ang{130}, \ang{140}, \ang{150}, \ang{170}\}\}$, always starting at $\theta=0$. The gripper moved $\SI{1}{\centi\meter}$ in each of the trial directions, and we checked if the contact point $B$ (as shown in Fig. \ref{fig:robust-overview}(a)) had separated. We verify contact separation by trying to insert a paper sheet between the horizontal plane and the peg. In all six trials, after the gripper motion, we could easily insert the paper sheet, which means contact point B had indeed separated. When the starting angle is $\varphi=\ang{38}$, the proposed method was able to plan a reorientation trajectory and could be successfully executed in the experiments. All these experiments are shown in the supplementary video.

The two test cases verify the predicted feasibility or infeasibility of the proposed in-hand motion model, which supports its accuracy.

\subsection{Effectiveness of Robust Planning}

\begin{figure}[!t]
\centering
\begin{subfigure}{0.23\textwidth}
    \centering
    \includegraphics[width=\textwidth, trim=18cm 0cm 12cm 1cm, clip=true]{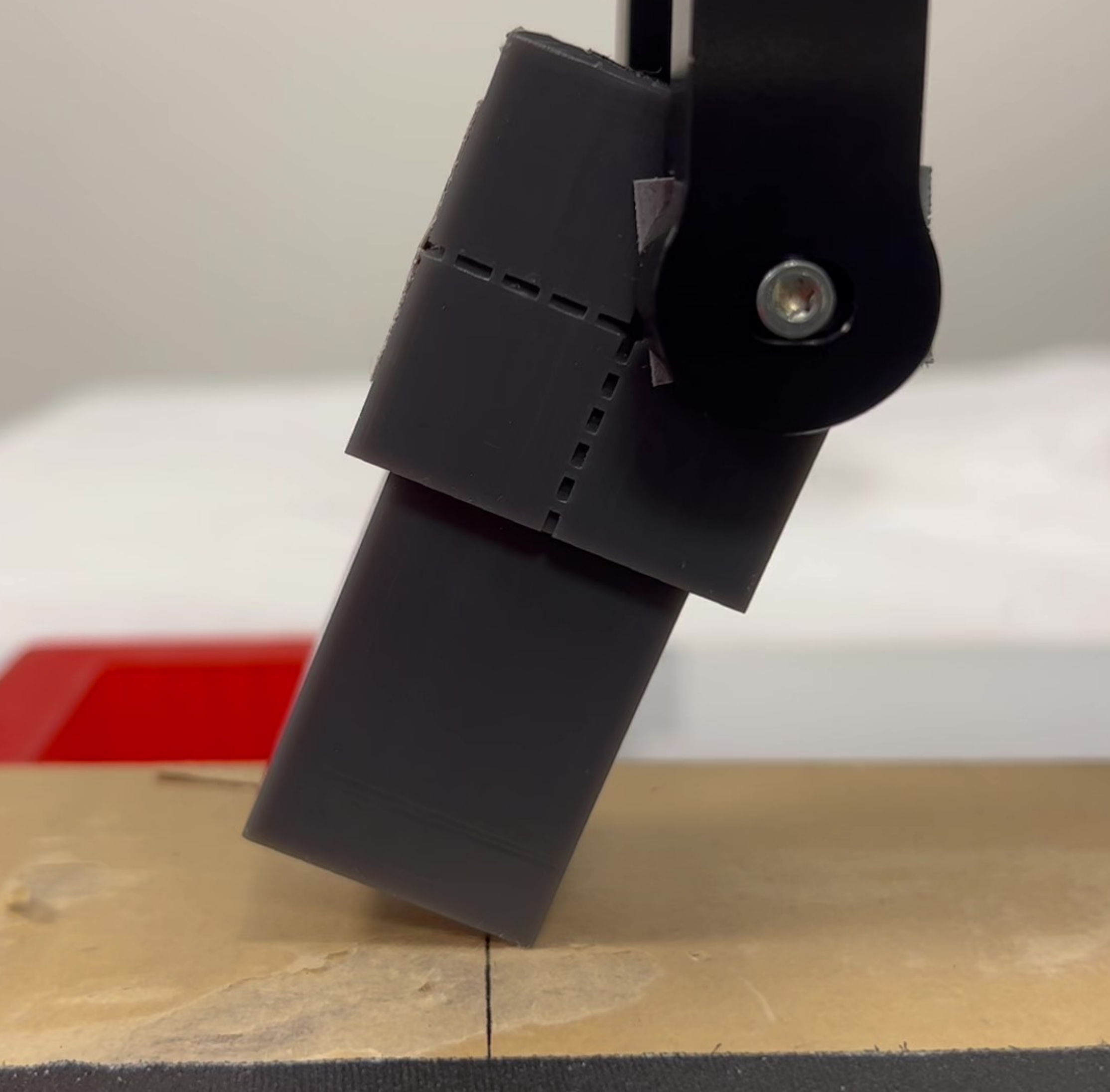}
    \caption{Na\"ive planning}
\end{subfigure}
\begin{subfigure}{0.23\textwidth}
    \centering
    \includegraphics[width=\textwidth, trim=18cm 1cm 12cm 0cm, clip=true]{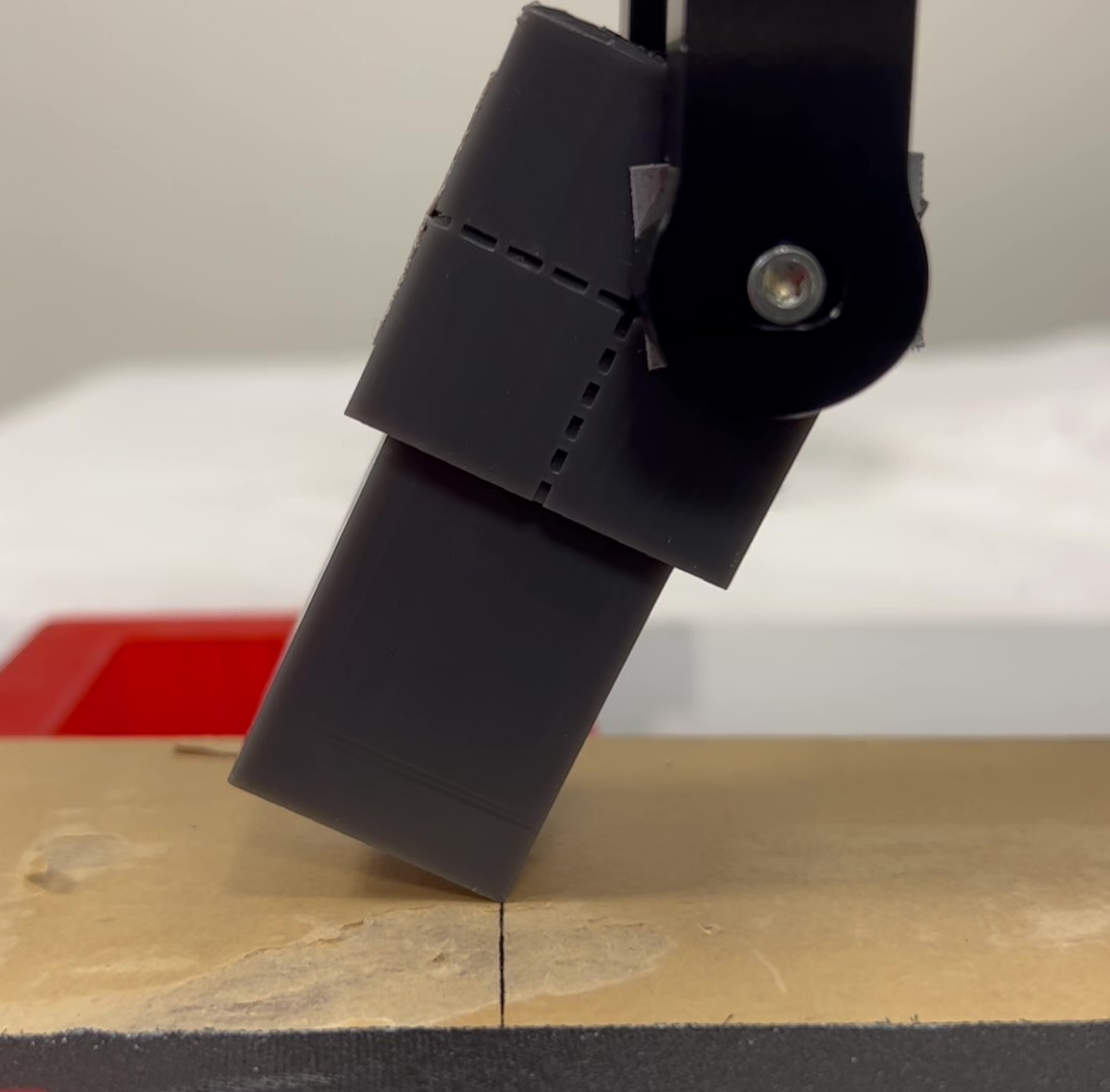}
    \caption{Robust planning}
\end{subfigure}
\caption{Snapshots of robust and na\"ive planning on single point pivoting under settings (x, \SI{-2}{\milli\meter}) and (y, +\SI{2}{\milli\meter}). The robust planning ensured the contact line does not slip as desired while na\"ive planning showed observable slip.}
\vspace{-0.5em}
\label{fig:exp-single-slide}
\end{figure}

\begin{table}[!t]
\begin{center}
\begin{tabular}{|c||c|c|c|}\hline
\diagbox[width=7em]{Na\"ive}{Robust} & \makebox[5em]{x, \SI{-2}{\milli\meter}} & \makebox[5em]{x, Exact} & \makebox[5em]{x, \SI[retain-explicit-plus]{+2}{\milli\meter}}\\\hline\hline
\makebox[6em]{y, \SI{-2}{\milli\meter}} & \diagbox[width=7em]{\SI{4.6}{\milli\meter}}{\SI{1.7}{\milli\meter}} & \diagbox[width=7em]{\SI{2.4}{\milli\meter}}{NE} & \diagbox[width=7em]{NE}{NE} \\\hline
\makebox[6em]{y, Exact} & \diagbox[width=7em]{\SI{3.8}{\milli\meter}}{NE} & \diagbox[width=7em]{2.2 mm}{NE} & \diagbox[width=7em]{NE}{NE}\\\hline
\makebox[6em]{y, \SI[retain-explicit-plus]{+2}{\milli\meter}} & \diagbox[width=7em]{\SI{3.5}{\milli\meter}}{NE} & \diagbox[width=7em]{\SI{1.6}{\milli\meter}}{NE} & \diagbox[width=7em]{NE}{NE}\\\hline
\end{tabular}
\end{center}
\caption{Sliding distance of the contact point. NE stands for negligible, which means no slippery is visually recognizable.}
\vspace{-2em}
\label{table:exp-slide-dist}
\end{table}

This subsection elaborates on the effectiveness of the robust planning method by comparing with the na\"ive planning, where the action is selected from the na\"ive motion cone $\mathbf{\Theta}_g$ (as mentioned in Section \ref{sec:mechanism} and \ref{sec:robust}). The example in Fig. \ref{fig:model_single_pivot_env} is used as the first test case. We consider the uncertainty in the grasp center ($x_g$ and $y_g$) and the CoM position ($x_c$ and $y_c$). We assume the grasp center and CoM position have $\SI{3}{\milli\meter}$ uncertainty in both $x$ and $y$ direction and generated the gripper motion trajectory using the robust planning method. The benchmark trajectory is generated using the na\"ive planning method. We intentionally added $\pm\SI{2}{\milli\meter}$ displacement, respectively, to the $x$ and $y$ coordinates of the grasping point and compared the performance of robust and na\"ive planning. The results are summarized in table \ref{table:exp-slide-dist}. Robust planning was able to ensure sticking contact in almost all cases, while na\"ive planning showed detectable contact point slip. The robust planning also had a $\SI{1.7}{\milli\meter}$ slip when both the $x$ and $y$ coordinates of the grasp center have $\SI{-2}{\milli\meter}$ error. This is most likely because the contact point fell out of the planned uncertainty range in this setting. Snapshots of na\"ive and robust planning are shown in Fig. \ref{fig:exp-single-slide}.

\begin{figure}
\centering
\begin{subfigure}{0.15\textwidth}
    \includegraphics[width=\textwidth, trim=14cm 2cm 12cm 2cm, clip=true]{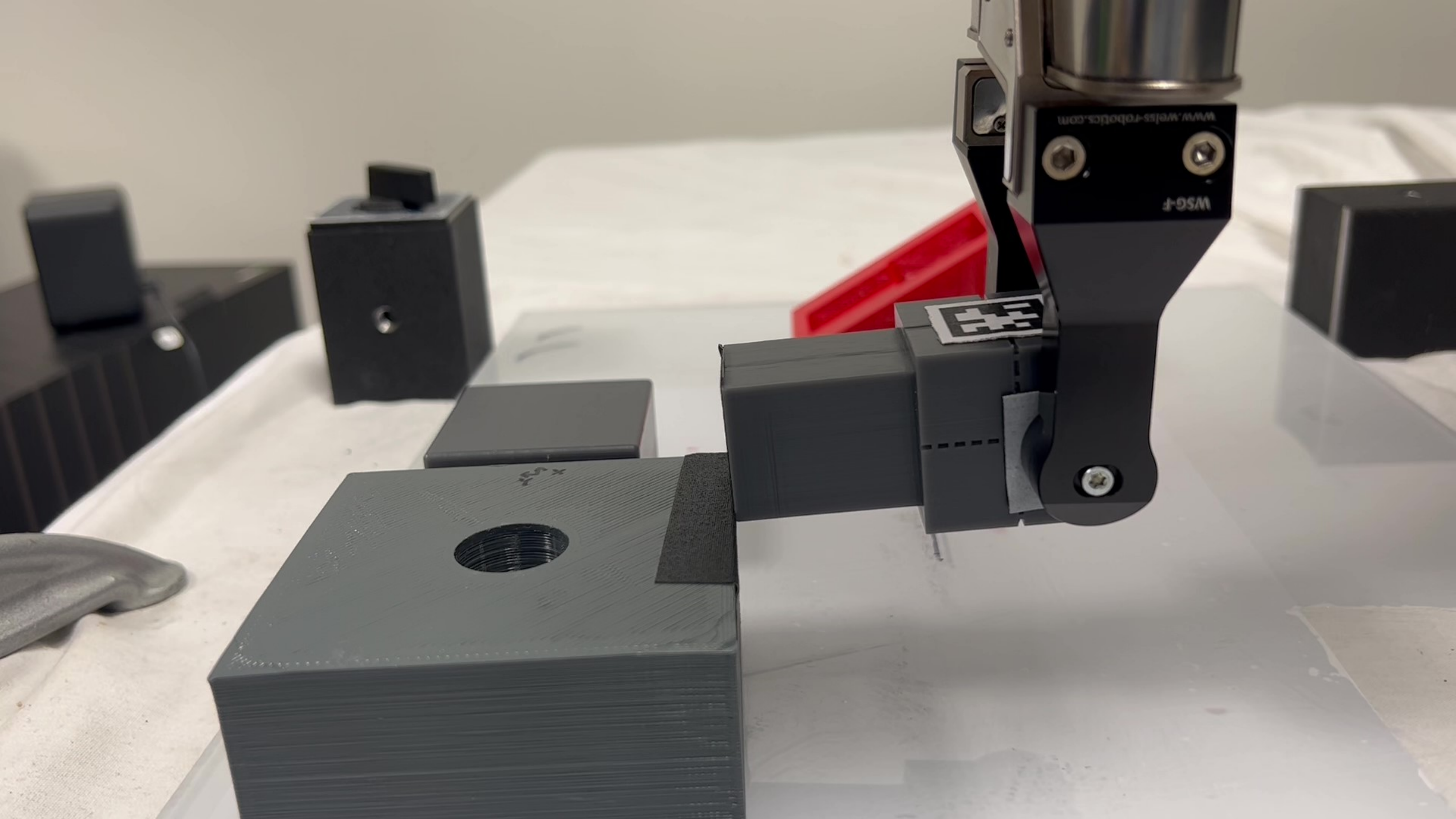}
    \caption{Starting Pose}
\end{subfigure}
\begin{subfigure}{0.15\textwidth}
    \includegraphics[width=\textwidth, trim=14cm 2cm 12cm 2cm, clip=true]{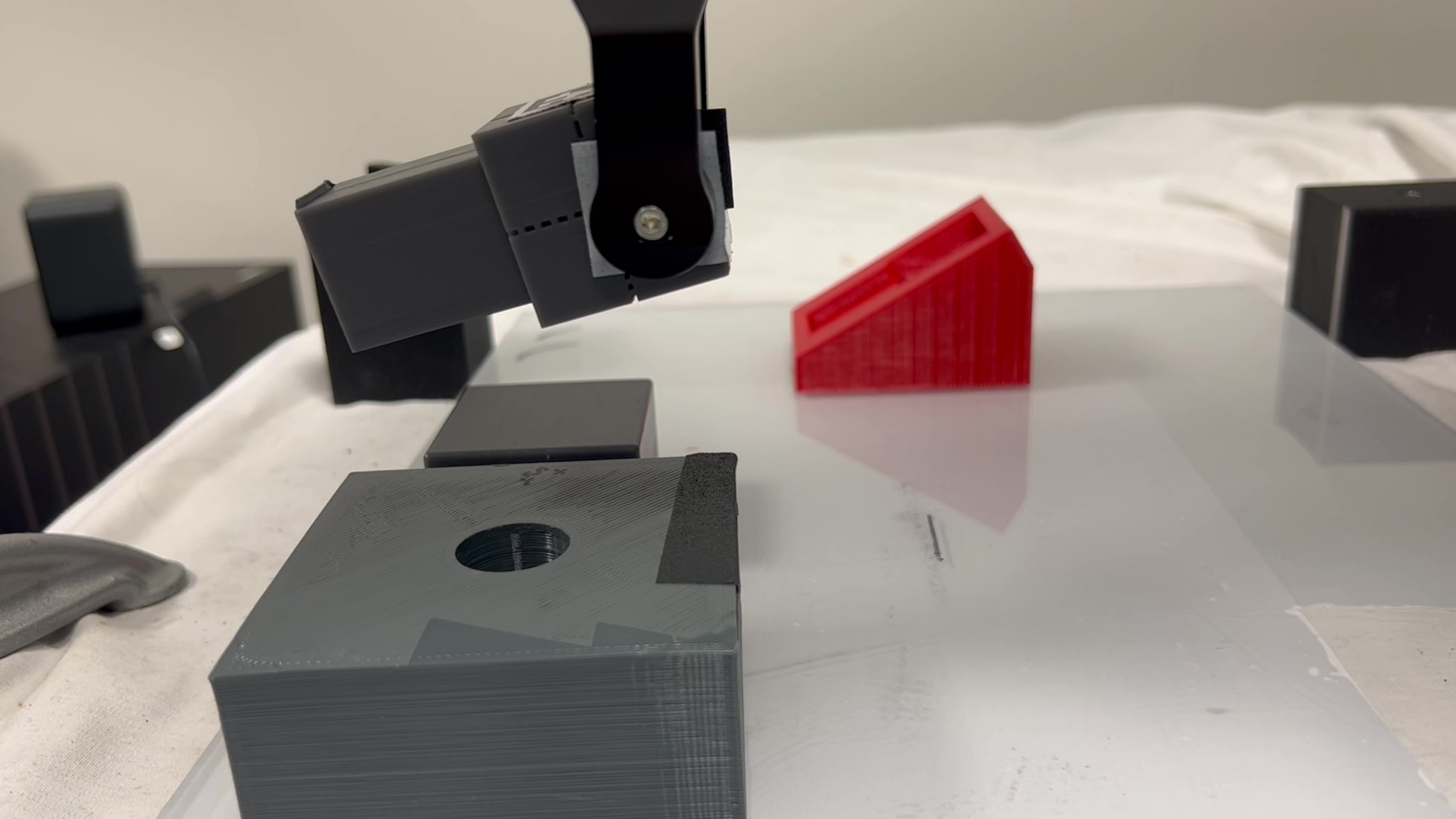}
    \caption{Na\"ive planning}
\end{subfigure}
\begin{subfigure}{0.15\textwidth}
    \includegraphics[width=\textwidth, trim=14cm 2cm 12cm 2cm, clip=true]{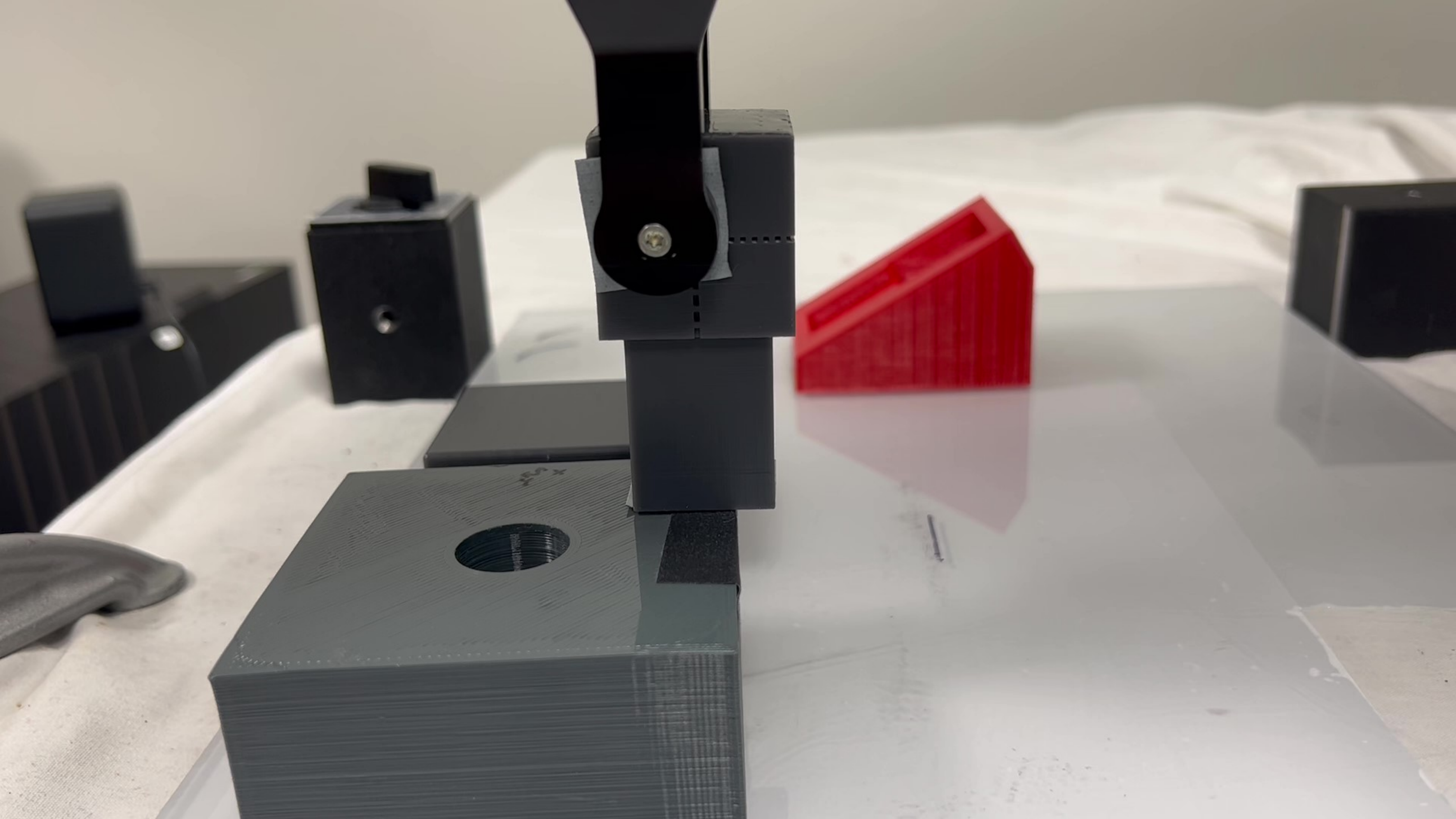}
    \caption{Robust planning}
\end{subfigure}
\caption{Snapshots of pivoting using an environmental edge. Na\"ive planning tends to lose external contact and fail the pivoting, while robust planning can maintain external contact and complete pivoting.}
\vspace{-0.5 em}
\label{fig:exp-motion-c}
\end{figure}

The second test case is when the robot needs to re-orient the object from horizontal to vertical pose while maintaining contact with an edge of a cuboid, as shown in Fig. \ref{fig:exp-motion-c}. Failure is declared if the object loses contact with its environment during execution. 
We executed the robust and na\"ive trajectory $10$ times each. Robust planning succeeded in all $10$ trials and na\"ive planning only marginally succeeded in two trials (meaning the peg was almost losing contact when reorientation completes).

\begin{figure}[t]
\centering
\begin{subfigure}{0.15\textwidth}
    \includegraphics[width=\textwidth, trim=9cm 0cm 10cm 3cm, clip=true]{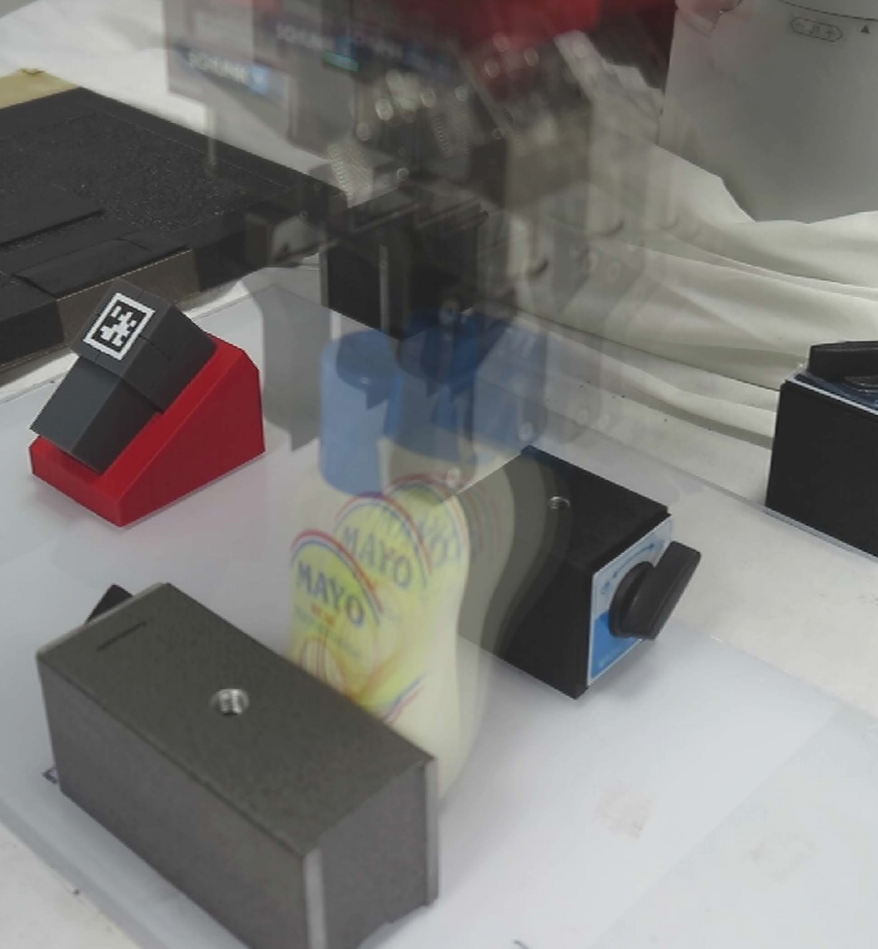}
    \caption{Mayo Bottle}
\end{subfigure}
\begin{subfigure}{0.15\textwidth}
    \includegraphics[width=\textwidth, trim=9cm 0cm 10cm 3cm, clip=true]{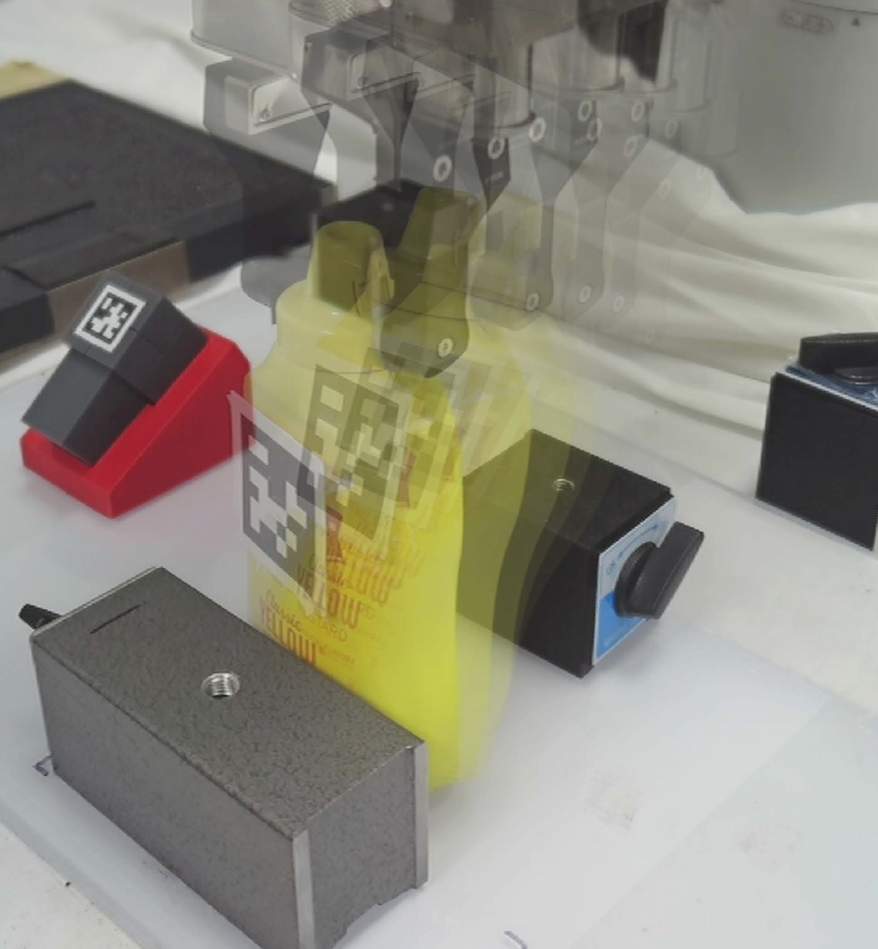}
    \caption{Mustard Bottle}
\end{subfigure}
\begin{subfigure}{0.15\textwidth}
    \includegraphics[width=\textwidth, trim=9cm 0cm 10cm 3cm, clip=true]{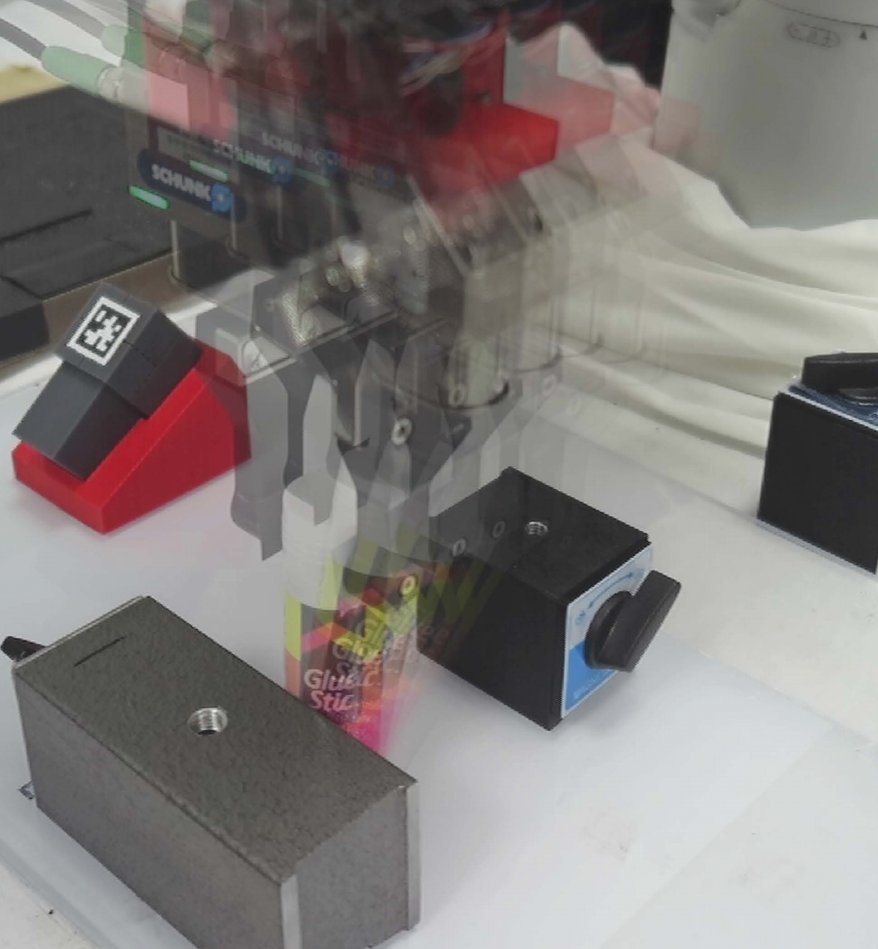}
    \caption{Glue Bottle}
\end{subfigure}
\caption{Snapshots of two point pivoting on three different daily objects.}
\label{fig:exp-daily}
\vspace{-2em}
\end{figure}

Thus, we show that the robust planning method can effectively maintain the contact mode against parametric uncertainties. We also successfully implemented robust planning on three additional daily objects as shown in Fig. \ref{fig:exp-daily}. Only a rough estimate of the contact point and CoM position was given to the system. This shows that our proposed manipulation framework is generalizable to various object shapes.

\section{Conclusion and Future Work}
Generalizable manipulation remains one of the biggest challenges in robotics community due to the challenges imposed by contact-rich interactions during manipulation. Planning robust behavior of a robot in the presence of uncertain contact parameters is very challenging and is not well explored in literature.
This paper proposed a robust planning method for in-hand manipulation task while maintaining extrinsic contacts. Given a desired contact mode and uncertainty range of each parameter, it first uses an in-gripper mechanics model to compute the motion cone for each combination of vertex parameters of the uncertainty range and takes their intersection to obtain the robust motion cone. Real world experiments were conducted to verify the accuracy of the in-gripper mechanics model. The comparison between the proposed robust planning and na\"ive planning also elaborated on the effectiveness of the robust tuning framework for maintaining contact mode under parametric uncertainties. Since the proposed method gives a robust motion cone for each contact mode and can provide forward dynamics, it can naturally combine with search-based algorithms such as rapidly exploring random tree search, which forms an interesting application and future work.

There are nevertheless limitations of our proposed method. In the robust planning process, the robust motion cone $\bar{\mathbf{\Theta}}_g$ may sometimes become an empty set for longer horizons and larger uncertainty range. As such, combining with tactile feedback methods~\cite{shirai2023tactile, kim2023simultaneous, ota2023tactile} becomes an important future work. 

\section*{Acknowledgment}

The authors thank Xinghao Zhu and Linfeng Sun for the fruitful discussion and constructive advice.

\bibliographystyle{IEEEtran}
\bibliography{references.bib}

\end{document}